\documentclass[10pt,twocolumn,letterpaper]{article}

\usepackage{float}
\usepackage{iccv}
\usepackage{times}
\usepackage{epsfig}
\usepackage{graphicx}
\usepackage{amsmath}
\usepackage{amssymb}
\usepackage{multirow}
\usepackage{bbding}

\usepackage[breaklinks=true,bookmarks=false]{hyperref}

\iccvfinalcopy 

\def\iccvPaperID{****} 

\ificcvfinal\fi

\begin{document}

\title{ An Attractor-Guided Neural Networks for
 Skeleton-Based Human Motion Prediction}

\author{Pengxiang Ding\\

\and
Junying Wang\\

\and
Jianqin Yin\\

}

\maketitle
\ificcvfinal\thispagestyle{empty}\fi

\begin{abstract}

Joint relation modeling is a curial component in human motion prediction. Most existing methods tend to design skeletal-based graphs to build the relations among joints, where local interactions between joint pairs are well learned. However, the global coordination of all joints, which reflects human motion's balance property, is usually weakened because it is learned from part to whole progressively and asynchronously. Thus, the final predicted motions are sometimes unnatural. To tackle this issue, we learn a medium, called balance attractor (BA), from the spatiotemporal features of motion to characterize the global motion features, which is subsequently used to build new joint relations. Through the BA, all joints are related synchronously, and thus the global coordination of all joints can be better learned. Based on the BA, we propose our framework, referred to Attractor-Guided Neural Network, mainly including Attractor-Based Joint Relation Extractor (AJRE) and Multi-timescale Dynamics Extractor (MTDE). The AJRE mainly includes Global Coordination Extractor (GCE) and Local Interaction Extractor (LIE). The former presents the global coordination of all joints, and the latter encodes local interactions between joint pairs. The MTDE is designed to extract dynamic information from raw position information for effective prediction. Extensive experiments show that the proposed framework outperforms state-of-the-art methods in both short and long-term predictions in H3.6M, CMU-Mocap, and 3DPW.

\end{abstract}

\section{Introduction}

3D skeleton-based human motion prediction aims to generate future skeleton sequences given past observed ones. This technique can help machines better understand human intention and have a broad prospect in scenarios such as human-robot interaction \cite{03,06,07}, autonomous driving \cite{01}, and pedestrian tracking \cite{02,05}.

\begin{figure}[htb] 
\begin{center}
   \includegraphics[width=0.45\textwidth]{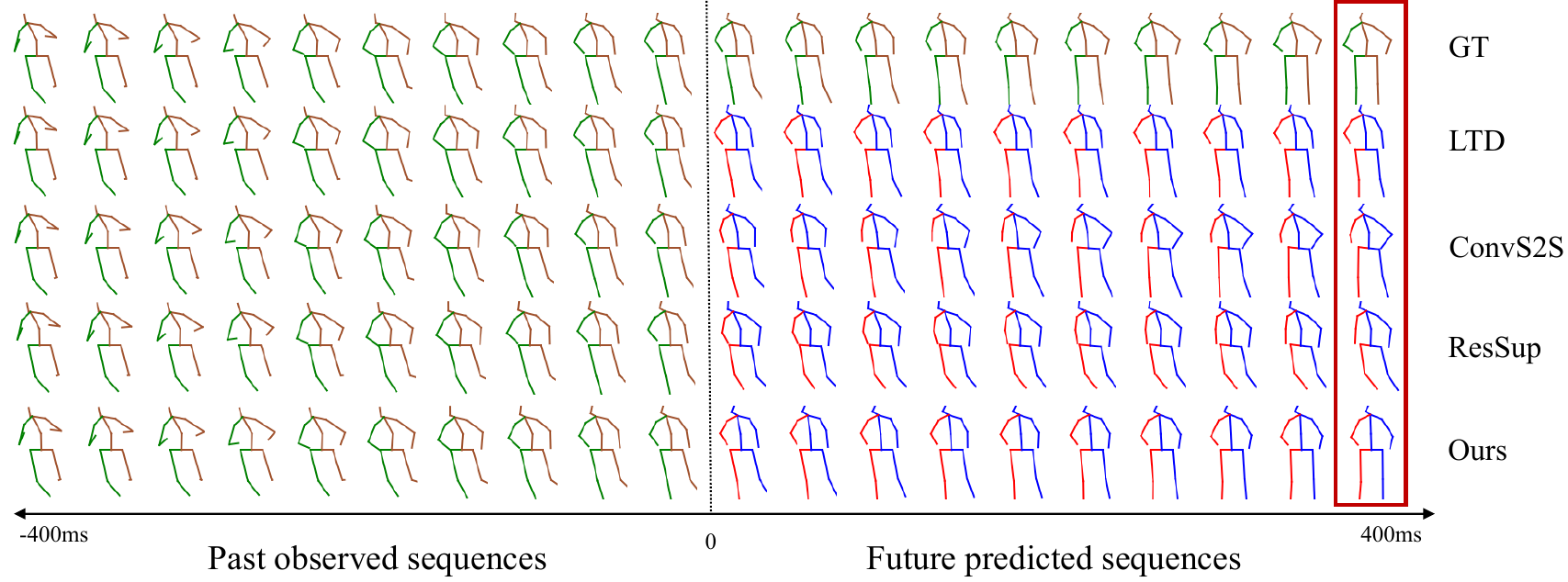} 
\end{center}
\caption{Qualitative results of short-term predictions of motion "discussion" on H3.6M. From top to bottom, we show the ground truth, the results of LTD \cite{08}, ConvS2S \cite{19}, ResSup \cite{17} and our approach. Compared with the result of our approach, the predicted motions of other works have the same problem: the limbs are uncoordinated which makes the predicted motion appear unnatural.}
\label{f_vis_all}
\end{figure}

Joint relation modeling is essential for motion prediction. Prior works mainly relied on graphs to model joint relation combined with neural networks, such as RNN \cite{16,17,18,21,22}, CNN \cite{19,20}, GCN \cite{23,33}. Most graphs are designed according to the kinematic structure of the human to extract motion features. Though they are effective, it is hard for them to learn the relations between spatial separated joint pairs directly. Recently, dynamic graphs were developed by \cite{08,34} to model the relations explicitly. Thus, the local interactions between joint pairs can be learned adequately. However, there still exists one drawback. The global coordination of all joints, which contributes to the balance of human motion, is not well learned. It is mainly because the global motion features are usually extracted by fusing different body components' local features. In this process, all joints' global relations are learned progressively and asynchronously, and thus the relations are usually weakened. It sometimes makes the predicted motion appears unnatural, e.g., the limbs are uncoordinated, as is shown in Figure \ref{f_vis_all}.

In this paper, we aim to learn the global coordination of all joints. To this end, we learn a balance attractor (BA) to act as the medium to build new relations of all joints indirectly. Specifically, the BA is learned by calculating dynamic weighted aggregation of single joint feature. Then we calculate the difference between the BA and each joint feature. Finally, the resulting new joint features are used to calculate joints similarities to generate final joint relations. In this way, all joints are related indirectly but synchronously through the BA. Meanwhile, because the new joint relations encode global motion features, the global coordination of all joints can be better learned.

\begin{figure*}[htb] 
\begin{center}
   \includegraphics[width=0.9\textwidth]{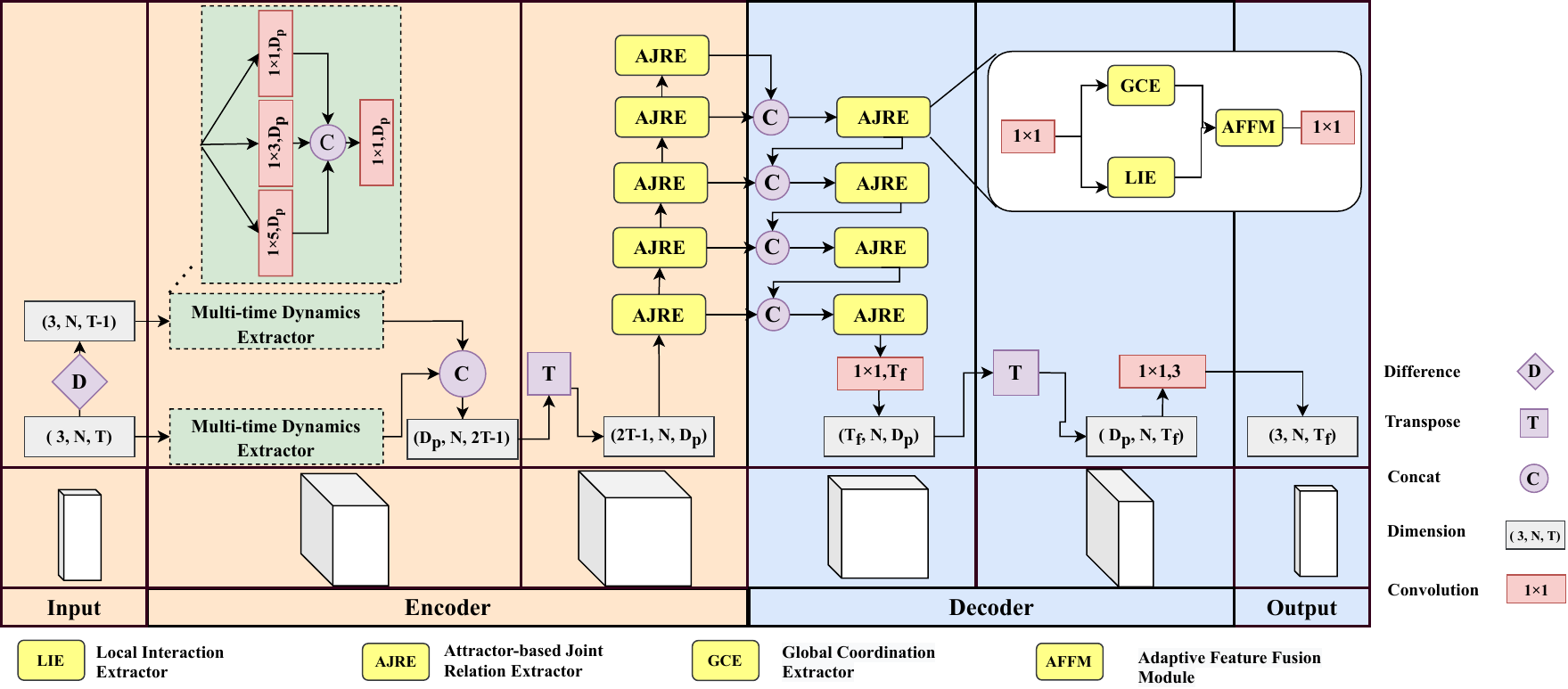} 
\end{center}
\caption{The framework of proposed Attractor-Guided Neural Network. In the encoder, a MTDE module is used to extract dynamics features of motion. The AJRE module is adopted to encode the global coordination of all joints and local interactions between joint pairs through GCE and LIE, respectively. AFFM is introduced to fuse features according to channel-wise attention. The whole AJRE is built based on the bottleneck architecture of ResNet \cite{27}.In the decoder, skipped connections are used to offer fine-grained information inspired by U-Net \cite{28}. The two $1\times 1$ convolutions are successively used to conduct space transformation to get final prediction results.}

\label{f_all}
\end{figure*}

Additionally, enriching dynamic representation of raw input data is also beneficial for effective prediction. As is well known, the raw skeleton sequences only include each joint's position information of different time steps, which are not sufficient to convey the dynamics property of motion. Previous works \cite{10,31} tended to introduce a two-stream architecture for extra velocity information. \cite{10,11} enlarge the time horizon by taking three neighbor frames into account. But it still ignores other dynamic information like accelerated speed, which is not limited to fixed timescales. Therefore, we extract the features among frames with multiple timescales to get enriching dynamic representation from raw 3D coordinates.

Based on the above two aspects, we present our framework referred to as Attractor-Guided Neural Network. Given observed motion sequences, we first learn an enriching dynamic representation from raw position information adaptively through Multi-timescale Dynamics Extractor (MTDE). Next, we introduce the Attractor-Based Joint Relation Extractor (AJRE), including a Local Interaction Extractor (LIE), a Global Coordination Extractor (GCE), and an Adaptive Feature Fusing Module. The LIE is used to encode the local interactions between joint pairs, and the GCE is designed to present the global coordination of all joints. The above different joint relations are adaptively aggregated in the Adaptive Feature Fusing module. 

The main contributions of this paper are summarized as follows.
1. We propose a novel joint relation modeling module, AJRE, mainly including GCE and LIE. GCE is proposed to model the global coordination of all joints, encoding the balance property of human motion. LIE is presented to mine the local interactions between joint pairs.
2. We also put forward an MTDE module to extract enriching dynamic
information from raw input data for effective prediction.
3. Our proposed Attractor-Guided Neural Network outperforms most state-of-the-art methods for short and long-term motion prediction on three standard benchmark datasets: H3.6M, CMU-Mocap, and 3DPW.

\section{Related work}

Skeleton-based motion prediction has attracted increasing attention recently. Recent works using neural networks \cite{08,16,17,18,19,20,21,22,23,34,33,32,38,35} have significantly outperformed traditional approaches \cite{12,13}.

\textbf{Human motion prediction.} RNNs\cite{16,17,18} are first used to predict human motion for their ability on sequence modeling. The first attempt was made by Fragkiadaki et al. \cite{16}, who proposed an Encoder-Recurrent-Decoder (ERD) model to combine encoder and decoder with recurrent layers. They encode the skeleton in each frame to a feature vector and built temporal correlation recursively. Julieta et al. \cite{17} introduced a residual architecture to predict velocities and achieved better performance. However, these works all suffer from discontinuities between the observed poses and the predicted future ones. Though Gui et al. \cite{18} proposed to generate a smooth and realistic sequence through adversarial training, it is hard to alleviate error-accumulation in a long-time horizon inherent to the RNNs scheme. A feedforward network was widely adopted to help alleviate those above questions because their prediction was not recursive and thus could avoid error-accumulation. Li et al. \cite{19} introduced a convolutional sequence-to-sequence model that encodes the skeleton sequence as a matrix whose columns represent the pose at every time step. However, their spatiotemporal modeling is still limited by the convolutional filters' size. Recently, \cite{08,20} were proposed to consider global spatial and temporal features simultaneously. They all transform temporal space to trajectory space to take the global temporal information into account. It contributes to capturing richer temporal correlation and thus achieved state-of-the-art results. In this paper, we follow this scheme but use different methods to model global spatial correlation.

\textbf{Joint relation modeling.} 
Previous work mainly focused on skeletal constraints to model correlations between joints. Jain et al. \cite{21} first introduced a Structural-RNN model to explicitly model structural information relying on high-level spatiotemporal graphs. However, the graph is designed according to kinetic structure and is not flexible for different motions. Recently, some dynamic graph structures \cite{08,32,23,36} were developed to model more flexible joint relations. Mao et al. \cite{08} used an adaptive graph to model motion, but it is still unreliable because the graph is initialized randomly without structure prior. Cui et al. \cite{32}further combined kinematic structure with dynamic graph structure.
Li et al. \cite{23} used stacked GCNs to build the interaction of different scales structure in each layer to model the correlation of both neighbor and distant joints. However, there still exists a problem in existing methods: the global coordination of all joints, which reflects the balance property of human motion, is usually weakened because they are learned from part to whole progressively and asynchronously. Therefore, in this paper, we aim to encode the global coordination of all joints. Based on this intuition, we propose an Attractor-Based Joint Relation Extractor (AJRE) to better leverage global coordination of all joints combined. Among the module, local interactions between joint pairs are also included as auxiliary information.

\textbf{Dynamic representation of Skeleton sequence.} 
Considering the raw skeleton sequence only represents each joint's position information at each time step, which is not sufficient to convey the dynamic property of motion. Many attempts \cite{10,11,31} proposed to extract enriching dynamic representation from raw data. They relied on two-stream architecture to introduce velocity information. A drawback of them is that they only extract the dynamics from neighbor frames. Though Li et al. \cite{10} enlarged the time horizon by convolution operation, it is still insufficient because dynamics exist in different timescales. Therefore, in this paper, we extract the dynamic features among frames through multiple timescales convolution and fuse them for enriching dynamic representation from raw 3D coordinates.

\section{Our Method}

The proposed balance attractor guided framework, AGN, models human motion from a new perspective. It mainly includes two components, MTDE and AJRE. MTDE extracts multi-time scale temporal information to obtain rich features for motion prediction. AJRE mines the balance attractor based dynamics from the multi-time scale input to model the spatiotemporal evolution of human motion. Finally, the two $1\times 1$ convolutions are successively used to conduct dimension reduction to get final predictions.

\subsection{Problem formulation}

We denote the historical 3D skeleton-based poses as ${X}_{1:T}=\left[x^1,\cdots,x^T\right]\in\mathbb{R}^{N\times T\times D}$ and future poses as ${X}_{T+1:T_f} = \left[x^{T+1},\cdots,x^{T_f}\right]\in\mathbb{R}^{N\times (T_f-T)\times D}$, where $x^t\in\mathbb{R}^{N\times D}$ represents the 3D pose at time $t$ with $N$ joints. The $D$ depicts the dimension of joint coordinates. Our goal is to generate predicted poses, ${\hat{X}}_{T+1:T_f}=AGN(X_{1:T})$ through proposed framework $AGN$.

\subsection{Multi-timescale Dynamics Extractor (MTDE)}

Dynamics is a important motion property to represent the patterns of current motion and is used to anticipate future motion trends. Many previous works utilize two-stream architecture to offer different modality inputs like velocity related to motion. While it makes sense, it is still not suitable for all motions because the length of dynamics in different motions varies. Thus, most of the previous works are incapable of getting efficient dynamic representation of motion. In this part, we conduct a combination of different time scales motion dynamics to coordinate with two-stream architecture to address this issue. And more fine-grained dynamic information can be achieved in our proposed Multi-timescale Dynamics Extractor.

The architecture is shown in Figure \ref{f_all}. We take two-stream architecture: one path is raw input with the size of $[D, N, T]$ and another path is the difference between adjacent frames in raw input with the size of $[D, N, T-1]$ representing the velocity of raw input. Both paths are connected with a feature extractor which encodes dynamics through three different time scales. Especially, we model dynamics of each joint separately to avoid the interference of other joints. For motion prediction, it is beneficial to enable the model to extract a richer representation of a single joint before building the correlation between joints.

We here take $X_{1:T}$ as an example. Given the input $X_{1:T}$, we first use different $1\times k_i$ temporal convolutions ${\rm conv}_{k_i}^p$ with different timescale $k_i$ to generate new dynamic features. Formally,
\begin{equation}
D_{k_i}^p={\rm conv}_{k_i}^p\ast X_{1:T}\ ,\ D_{k_i}\in\mathbb{R}^{N\times T\times D_p}
\label{eq_2}
\end{equation}
where $i\in [1,2,3]$, $\ast$ indicates the convolution operation and $D_p$ is the size of new channel.

Considering different $D_{k_i}^p$ contains different dynamic features of motion, we concatenate them along the channel. This operation enables the model to capture coarse and subtle detailed dynamics simultaneously. Meanwhile, here we also use a $1\times1$ convolution ${\rm conv}_{red}^p$ to reduce feature channels for efficiency. Formally,
\begin{equation}
D_{concat}^p=\left[D_{k_1}^p;D_{k_2}^p;D_{k_3}^p\right],D_{concat}^p\in\mathbb{R}^{N\times T\times3D_p}
\label{eq_3_1}
\end{equation}
\begin{equation}
D_{red}^p={conv}_{red}^p\ast D_{concat}^p,D_{red}^p\in\mathbb{R}^{N\times T\times3D_p}
\label{eq_3_2}
\end{equation}
where $[\ ;]$ represents the concatenation along the channel.

Similar to $X_{1:T}$, we also extract the dynamics for $V_{1:T-1}$ with the same process to get the representation $D_{red}^V$. Specifically, $V_{1:T-1}=X_{2:T}-X_{1:T-1}$ is calculated by makeing differences between adjacent frames of $X_{1:T}$. To make use of different features, we synthesize them along temporal dimensions to get dynamic representation.
\begin{equation}
D_{red}^{all}\ =\ D_{red}^p\ \oplus D_{red}^v,D_{red}^{all}\in\mathbb{R}^{N\times \left(2T-1\right)\times D_p}
\label{eq_4}
\end{equation}
where $\oplus$ represents the concatenation along temporal dimension.

\subsection{Attractor-Based Joint Relation Extractor (AJRE)}

\begin{figure}[H] 
\begin{center}
   \includegraphics[width=0.4\textwidth]{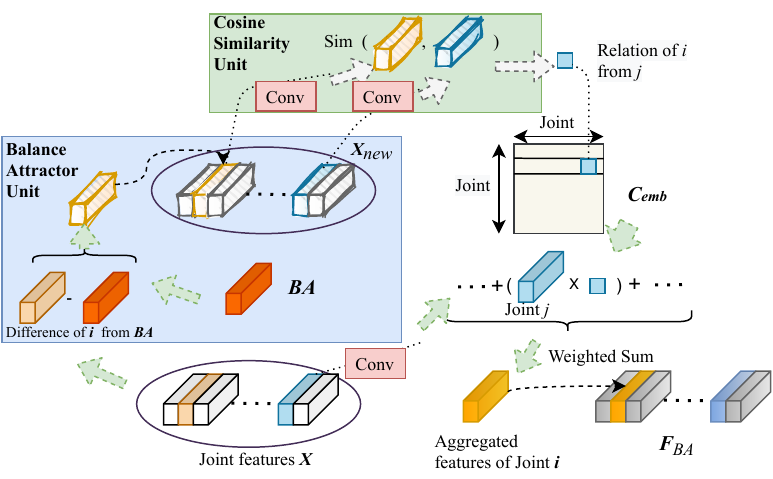} 
\end{center}
\caption{The overall process of GCE.}
\label{f_GCE2}
\end{figure}

The AJRE is used to exploit more prosperous joint relations of motion to help effective modeling. We thus propose Global Coordination Extractor (GCE) and Local Interaction Extractor (LIE) to separately model global coordination of all joints and local interactions between joint pairs. The Adaptive Feature Fusion module (AFFM) is introduced to fuse features according to channel-wise attention to improve the flexibility of joint relation modeling.

\subsubsection{Global Coordination Extractor (GCE)}

Global coordination of all joints plays an essential role in human motion. It needs all joints to coordinate synchronously and controls the balance of the human body during motion. However, it is usually weakened in previous works because the global motion features are generally learned by fusing the local features of different body components asynchronously and progressively. To tackle this issue, we learn a medium to build new joint relations indirectly. Through the medium, all joints are related synchronously, and thus the global coordination of all joints can avoid being weakened and thus it can be better learned.

\begin{figure}[H] 
\begin{center}
\includegraphics[width=0.5\textwidth]{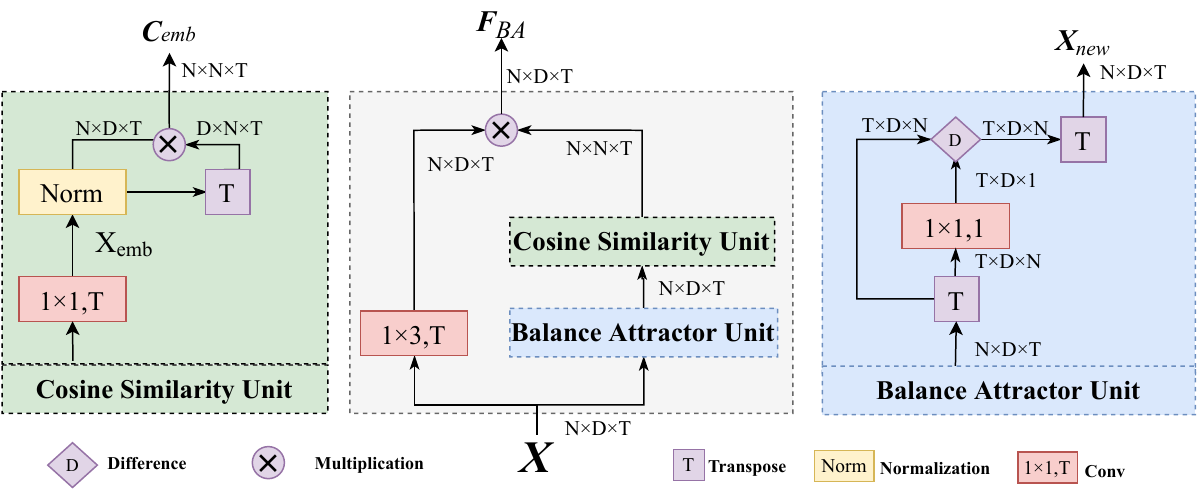} 
\end{center}
\caption{The implementations of GCE. The module has two paths in parallel. One path invloves pure $1\times3$ convolution. Another path contains serial Balance Attractor Unit and Cosine Similarity Unit.}
\label{f_GCE}
\end{figure}

As is shown in Figure \ref{f_GCE2}, we illustrate how to learn global coordination of all joints through the BA. In the Balance Attractor Unit, we first learn a medium called balance attractor (BA) by calculating all joints' aggregation to characterize the global motion features. We then calculate the difference between the BA and each joint feature to fuse the global motion features into each joint feature. In the Cosine Similarity Unit, we generate a new joint relation by measuring the similarities of joint pairs' new features. In this way, all joints can be related synchronously through this medium and thus can reflect the global coordination of all joints. The relation graph is subsequently to guide the motion feature extraction. It is noteworthy that we learn the BA in high dimensional space instead of in 3D space because the spatiotemporal features of motion in high dimensional space represent more dynamics. Besides, we name the medium as the balance attractor because it is used to model all joints' global coordination, which equals human motion's balance property.

More details are illustrated in Figure \ref{f_GCE}. In the Balance Attractor Unit, given input $X\in\mathbb{R}^{N\times D\times T}$, we first do the dimension transpose to get $X^{Tr}\in\mathbb{R}^{T\times D\times N}$. The channel size is set to $N$, which represents the number of joints, and the resulting feature map in each channel with the size of $[T\times N]$ represents the spatiotemporal features of each joint. Next, we here adopt a simple $1\times1$ convolution $conv_{BA}$ to learn the BA. Because the output of a convolution is the global response of the input channel, the BA represents all joints' comprehensive features and reflects the global motion features. This process is a dynamic weighted feature aggregation of $N$ joints features. The weight is learned by $conv_{BA}$ and adaptive to different motions. Formally,

\begin{equation}
BA=conv_{BA} \ast X^{Tr}, BA\in\mathbb{R}^{T\times D\times 1}
\label{eq_5}
\end{equation}
where $Tr$ represents the transformation between the joint dimension and the temporal dimension.

After getting a BA, it is used as a medium to build a new representation $X_{new}$ relative to BA for each joint indirectly through making differences. The purpose is to fuse the global motion features into each joint feature.

\begin{equation}
X_{new}= {(X^{Tr}- BA)}^{Tr}, X_{new} \in\mathbb{R}^{N\times D\times T}
\label{eq_6}
\end{equation}

We focus on building new relations of all joints through $X_{new}$ in the Cosine Similarity Unit. This step aims to encode the coordination of all joints into the relative joint relations graph. Specifically, We first use a $1\times1$ convolution ${conv}_{emb}$ to learn a embedding of ${\ X}_{new}$.
\begin{equation}
X_{emb}={\rm conv}_{emb} \ast X_{new},X_{emb}\in\mathbb{R}^{N\times D\times T}
\label{eq_7}
\end{equation} 

Next, we aim to calculate the relative relations of joints. The size of
one feature map of $X_{emb}$, of which each row represents the spatiotemporal features of one joint, is $[N \times D]$. Therefore, we can calculate the cosine similarity between all row vector pairs to illustrate the correlation between joint pairs. The reasons why we choose cosine similarity are: (1) this metric contains angle information that corresponds to the mutual influence between joints; (2) the value is limited into [-1,1], which avoids the violent variance.

Formally, we denote $\alpha_n{\in\mathbb{R}}^D$ as a row vector of each feature map at channel $t$, where $n=1,\ldots,N$. And then we can calculate the correlation matrix as:
\begin{equation}
C_t(\alpha_1,...,\alpha_n)=\left(\begin{matrix}c\left(\alpha_1,\alpha_1\right)&...&c\left(\alpha_1,\alpha_n\right)\\...&...&...\\c\left(\alpha_1,\alpha_n\right)&...&c\left(\alpha_n,\alpha_n\right)\\\end{matrix}\right)
\label{eq_8}
\end{equation} 

\begin{equation}
c\left(\alpha_i,\alpha_j\right)=\frac{\left<\alpha_i,\alpha_j\right>}{\left|\alpha_i\right|\left|\alpha_j\right|},\ i,j=1,...,N
\label{eq_9}
\end{equation}

where $c\left(\alpha_i,\alpha_j\right)\in[-1,1]$ represents similarity of $\alpha_i$ and $\alpha_j$, $C_t(\alpha_1,...,\alpha_n){\in\mathbb{R}}^{N\times N}$ denotes the correlation between all joints.

Notably, we calculate the correlation matrix on each channel because each channel encodes specific spatiotemporal features and should focus on different correlations compared with other channels. Therefore, we can get the correlation matrix of all channels:
$$C_{emb}=[C_1,...,C_T], C_{emb}\in\mathbb{R}^{N\times N\times T}$$. 

The last step is to calculate the aggregated features according to the joint relation $C_{emb}$. Specifically, $1\times 3$ convolution $conv_{intra}$ is used to extract intra-joint features and then combine with the guidance of $C_{emb}$ to get the final features $F_{BA}$.

\begin{equation}
F_{BA}=C_{emb}\odot\left(conv_{intra}\ast X\right),F_{BA}\in\mathbb{R}^{N\times D\times T}
\label{eq_10}
\end{equation}
where $\odot$ represents channel-wise multiplication.

\begin{table*}[!t] 
\caption{Short-term prediction on H$3.6$M. Where ``ms'' denotes ``milliseconds''.}
\scriptsize
\begin{center}
\begin{tabular}{c|cccc|cccc|cccc|cccc}
\hline
motion & \multicolumn{4}{c}{Walking} & \multicolumn{4}{c}{Eating}& \multicolumn{4}{c}{Smoking} & \multicolumn{4}{c}{Discussion}\\
\hline
time(ms)&80&160&320&400&80&160&320&400&80&160&320&400&80&160&320&400 \\
\hline
ResSup \cite{17} &23.8 &40.4& 62.9& 70.9& 17.6& 34.7& 71.9& 87.7& 19.7& 36.6& 61.8& 73.9&31.7& 61.3& 96.0& 103.5 \\
ConvS2S \cite{19} &17.1 &31.2&53.8&61.5&13.7&25.9&52.5&63.3&11.1&21.0&33.4&38.3&18.9&39.3&67.7&75.7\\
LTD \cite{08}&8.9 &15.7&29.2& 33.4& 8.8& 18.9& 39.4& 47.2& 7.8& 14.9& 25.3&{28.7}& 9.8& 22.1&{39.6} &{\bf44.1} \\

LPJP \cite{32} &7.9 &14.5&29.1&34.5&8.4&18.1&37.4&45.3&6.8&{13.2}&{24.1}&{\bf27.5}&8.3&{21.7}&43.9&48.0\\

TrajCNN \cite{20} &8.2 &14.9&30.0&35.4&8.5&18.4&37.0&44.8&6.3&{\bf12.8}&{\bf23.7}&{27.8}&7.5&{\bf20.0}&41.3&47.8\\

\hline
 Ours&{\bf 7.2} & {\bf 13.7} &{\bf25.6}& {\bf31.0}& {\bf 7.7} &{\bf 16.7}&{\bf 35.8}&{\bf 44.2} &{\bf 6.3}&{ 13.3} &{24.5}&29.7&{\bf 7.5} &{20.3}&{\bf 38.7}&{44.7}\\

\hline
\hline
motion & \multicolumn{4}{c}{Direction} & \multicolumn{4}{c}{Greeting}& \multicolumn{4}{c}{Phoning} & \multicolumn{4}{c}{Posing}\\
\hline
time(ms)&80&160&320&400&80&160&320&400&80&160&320&400&80&160&320&400 \\
\hline
ResSup \cite{17} & 36.5 &56.4& 81.5& 97.3&37.9& 74.1& 139.0& 158.8 &25.6& 44.4& 74.0& 84.2& 27.9& 54.7& 131.3& 160.8 \\
ConvS2S \cite{19} & 22.0&37.2 &59.6& 73.4 &24.5 &46.2 &90.0& 103.1& 17.2& 29.7& 53.4 &61.3& 16.1& 35.6& 86.2& 105.6\\

LTD \cite{08}& 12.6 & 24.4&{48.2}&{ 58.4}& 14.5& 30.5& 74.2& 89.0 & 11.5& 20.2& 37.9& 43.2&9.4& 23.9& {66.2}&{ 82.9}\\

LPJP \cite{32} &11.1 &22.7&48.0&58.4&13.2&28.0&64.5&77.9&10.8&{19.6}&{37.6}&{46.8}&8.3&{22.8}&65.6&81.8\\

TrajCNN \cite{20} &9.7 &22.3&50.2&61.7&12.6&28.1&67.3&80.1&10.7&18.8&37.0&43.1&6.9&21.3&62.9&78.8\\

\hline
Ours&{\bf 9.3}&{\bf 21.1}&{\bf 45.0}&{\bf 55.0}&{\bf 11.2}&{\bf 23.9}&{\bf 63.4}&{\bf 79.6}& {\bf 10.2}&{\bf 18.5}& {\bf 34.3}&{\bf 38.5}  &{\bf 6.8}&{\bf 20.5}&{\bf 60.6}&{\bf 76.6}\\

\hline
\hline
motion & \multicolumn{4}{c}{Purchasing} & \multicolumn{4}{c}{Sitting}& \multicolumn{4}{c}{Sitting down} & \multicolumn{4}{c}{Taking photo}\\
\hline
time(ms)&80&160&320&400&80&160&320&400&80&160&320&400&80&160&320&400 \\
\hline
ResSup \cite{17} & 40.8& 71.8& 104.2& 109.8 &34.5& 69.9& 126.3 &141.6& 28.6& 55.3& 101.6& 118.9 &23.6 &47.4& 94.0& 112.7\\
ConvS2S \cite{19} & 29.4& 54.9& 82.2& 93.0 &19.8 &42.4& 77.0& 88.4& 17.1& 34.9& 66.3& 77.7& 14.0& 27.2& 53.8& 66.2\\
LTD \cite{08}& 19.6&38.5&{ 64.4}&{ 72.2}&10.7& 24.6& 50.6&62.0&11.4 &{ 27.6}& 56.4& 67.6& 6.8& 15.2& {38.2}&{49.6} \\

LPJP \cite{32} &18.5 &38.1&{\bf61.8}&{\bf69.6}&9.5&23.9&49.8&61.8&11.2&{29.9}&{59.8}&{68.4}&6.3&{14.5}&38.8&49.4\\

TrajCNN \cite{20}& 17.1 &{\bf 36.1}&{64.3}&{75.1}& 9.0& 22.0& 49.4& 62.6 & 10.7& 28.8& 55.1& 62.9&{\bf5.4}& {\bf13.4}& {\bf36.2}&{\bf47.0}\\

\hline

Ours&{\bf 17.1}&38.0&65.0&73.0& {\bf 7.8}&{\bf 19.9}&{\bf 44.9}&{\bf 56.4}& {\bf 9.2}&{\bf 23.7}&{\bf 47.7}&{\bf 59.4}&{5.6}&{14.3}&{37.6}&{48.9}\\
\hline
\hline
motion & \multicolumn{4}{c}{Waiting} & \multicolumn{4}{c}{Walking dog}& \multicolumn{4}{c}{Walking Together} & \multicolumn{4}{c}{Average}\\
\cline{1-17}
time(ms)&80&160&320&400&80&160&320&400&80&160&320&400&80&160&320&400 \\
\hline
ResSup \cite{17} & 29.5& 60.5& 119.9& 140.6& 60.5& 101.9& 160.8& 188.3& 23.5& 45.0& 71.3& 82.8& 30.8& 57.0& 99.8& 115.5\\
ConvS2S \cite{19} &17.9& 36.5& 74.9& 90.7& 40.6& 74.7& 116.6& 138.7& 15.0& 29.9& 54.3& 65.8& 19.6& 37.8& 68.1& 80.2\\
LTD \cite{08}& 9.5& 22.0& 57.5& 73.9& 32.2& 58.0& 102.2& 122.7 & 8.9& { 18.4}& 35.3& 44.3& 12.1& 25.0& 51.0& 61.3\\

LPJP \cite{32} &8.4 &21.5&53.9&69.8&22.9&50.4&100.8&119.8&8.7&{18.3}&{34.2}&{44.1}&10.7&{23.8}&50.0&60.2\\

TrajCNN \cite{20}& 8.2 & 21.0&{53.4}&{68.9}& 23.6& 52.0& 98.1& 116.9 & 8.5& 18.5& 33.9& 43.4&10.2& 23.2& {49.3}&{ 59.7}\\

\hline
Ours&{\bf 7.7}&{\bf 18.8}&{\bf 48.0}&{\bf 64.7} & {\bf 22.0}&{\bf 49.2}&{\bf 90.9}&{\bf 110.0}&  {\bf 7.8}&{\bf 17.3}&{\bf 32.1}&{\bf 43.3}&{\bf 9.6}&{\bf 22.0}&{\bf 46.2}&{\bf 57.0}\\

\hline

\end{tabular}

\end{center}
\label{r_h36mshort}
\vspace{-2.5em}
\end{table*}

\subsubsection{Local Interaction Extractor (LIE)}

\begin{figure}[htp] 
\begin{center}
\includegraphics[width=0.25\textwidth]{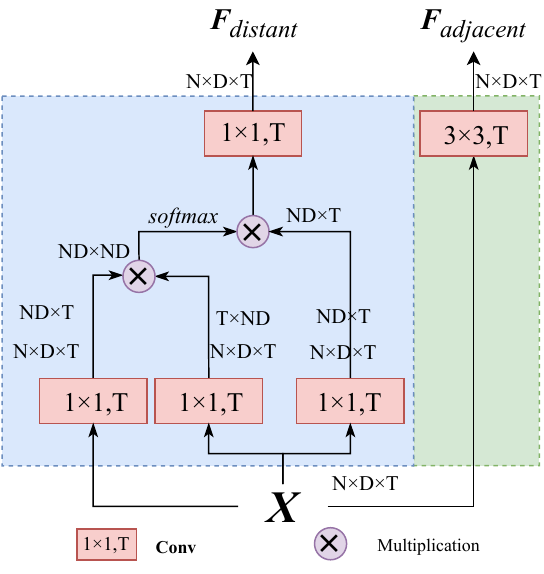} 
\end{center}
\caption{The implementations of Local Interaction Extractor (LIE). The left is the path using a non-local block without residual connection to learn the relations between distant joint pairs. The right is the the path with convolutions to learn the relations between adjacent joint pairs.}
\label{f_LIE}
\end{figure}

Local Interaction Extractor (LIE) is used to learn local interactions between joint pairs, including adjacent and distant joints. The local connection via bones brings spatial correlation for adjacent joints. For distant joints, some joints may have a strong correlation even if they are not directly connected, e.g., left hand and right hand are tightly correlated during ‘eating’. Therefore, these two relations are equally important for effective prediction.

As is shown in Figure \ref{f_LIE}, given an input $X$ which is the same as GCE, there exist two main paths to separately learn the relations between adjacent joint pairs and distant joint pairs. To learn the relations between adjacent joint pairs, a pure $3\times3$ convolution $conv_{adjacent}$ is adopted to extract spatiotemporal features between adjacent joint pairs. To learn the relations between distant joint pairs, the self-attention module Non-local \cite{30} is used to capture spatiotemporal features between adjacent joint pairs. The outputs can be described as follows. More details of this module are provided in the supplementary materials.
\begin{equation}
F_{adjacent}=conv_{adjacent}\ast X,\ F_{adjacent}\in\mathbb{R}^{N\times D\times T}
\label{eq_11}
\end{equation}
\begin{equation}
F_{distant}=Nonlocal(X),F_{distant}\in\mathbb{R}^{N\times D\times T}
\label{eq_12}
\end{equation}

\subsubsection{Adaptive Feature Fusing Module (AFFM)}
\begin{figure}[htp] 
\begin{center}
\includegraphics[width=0.4\textwidth]{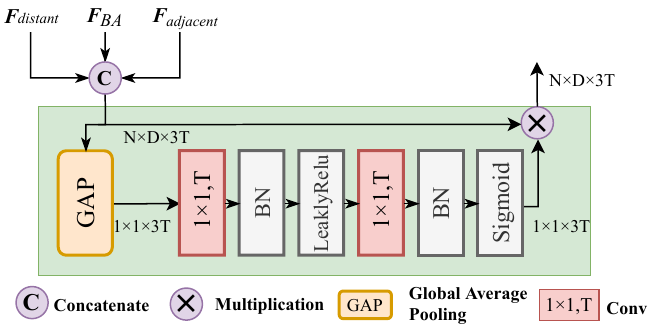} 
\end{center}
\caption{The implementations of Adaptive Feature Fusing Module (AFFM). The features learnt from previous block are fused with channel attention machanism.}
\label{f_AFFM}
\end{figure}
The different motions will have a respective preference for local interactions between joint pairs and global coordination of all joints. Here we adopt the channel attention mechanism to fuse features adaptively and reform more reliable representation. 

As is shown in Figure \ref{f_AFFM}, the global average pooling of the raw input represents the value of the feature map. After several operations of neural networks, we can get the importance ratio of each channel through the sigmoid function. Last we do channel-wise multiplication between ratio and raw input to reform features. More details of this module are provided in the supplementary materials.

\begin{table*}[!t] 
\vspace{0.5em}
\caption{Long-term prediction on H$3.6$M.}
\scriptsize
\begin{center}
\begin{tabular}{c|cc|cc|cc|cc|cc|cc|cc|cc}
\hline
motion & \multicolumn{2}{c}{Walking} & \multicolumn{2}{c}{Eating}& \multicolumn{2}{c}{Smoking} & \multicolumn{2}{c}{Discussion}
&\multicolumn{2}{c}{Directions} & \multicolumn{2}{c}{Greeting}& \multicolumn{2}{c}{Phoning} & \multicolumn{2}{c}{Posing}\\
\hline
time(ms)&560 &1000&560 &1000&560 &1000&560 &1000&560 &1000&560 &1000&560 &1000&560 &1000\\
\hline
LTD \cite{08}&42.2&51.3&{\bf56.5}&{\bf68.6}&32.3&60.5&{\bf70.4}&103.5&85.8&109.3&91.8&87.4&65.0&113.6&113.4&220.6\\
TrajCNN \cite{20}&37.9&46.4&59.2&71.5&32.7&58.7&75.4&{\bf103.0}&{\bf84.7}&104.2&91.4&{\bf84.3}&62.3&113.5&111.6&{\bf210.9}\\
\hline
Ours &{\bf35.5}&{\bf42.7}  &57.3&70.3 &{\bf30.9}&{\bf55.0} &74.3&105.7 &89.7&{\bf103.5} &{\bf91.1}&90.5  &{\bf59.1}&{\bf110.5} &{\bf107.3}&211.9 \\
\hline
\end{tabular}
\begin{tabular}{c|cc|cc|cc|cc|cc|cc|cc|cc}
\hline
motion & \multicolumn{2}{c}{Purchases} & \multicolumn{2}{c}{Sitting}& \multicolumn{2}{c}{Sitting down} & \multicolumn{2}{c}{Taking photo}
&\multicolumn{2}{c}{Waiting} & \multicolumn{2}{c}{Walking Dog}& \multicolumn{2}{c}{Walking Together} &\multicolumn{2}{c}{Average}\\
\cline{1-17}
time(ms)&560 &1000&560 &1000&560 &1000&560 &1000&560 &1000&560 &1000&560 &1000&560 &1000\\
\hline
LTD \cite{08}&94.3&130.4&79.6&114.9&82.6&140.1&68.9&87.1&100.9&167.6&136.6 &174.3&{\bf57.0}&85.0&78.5&114.3\\
TrajCNN \cite{20}&84.5&{\bf115.5}&81.0&116.3&79.8&{\bf123.8}&{\bf73.0}&{\bf86.6}&92.9&165.9&141.1&181.3&57.6&{\bf77.3}&77.7&110.6\\
\hline
Ours&{\bf82.1}&117.6&{\bf73.1}&{\bf105.1}&{\bf78.0}&126.1&75.9&88.9&{\bf85.9}&{\bf154.4}&{\bf130.2}&{\bf170.7}&57.1&82.2&{\bf75.1}&{\bf109.0}\\
\hline
\end{tabular}
\end{center}
\label{r_h36mlong}
\vspace{-2.0em}
\end{table*}

\subsection{Loss Function}
Following \cite{20,08}, we make use of the Mean Per Joint Position Error (MPJPE). In particular, for one training sample, loss is as follows:
\begin{equation}
L=\ \frac{1}{N\times\left(T_f-T\right)}\sum_{i=T+1}^{T_f}\sum_{j=1}^{N}{\parallel X_{i,j}-\ {\hat{X}}_{i,j}\parallel}_2
\label{eq_13}
\end{equation}
where ${\hat{X}}_{i,j}\in R^3$, representing the 3D coordinates of the $j_{th}$ joint of the $i_{th}$ human pose, is the predicted result and $X_{i,j}\in R^3$ is the ground truth.

\section{Experiments}

We evaluate our model on several benchmark motion capture (mocap) datasets, including Human3.6M (H3.6M) \cite{24}, the CMU mocap dataset, and the 3DPW dataset \cite{25}. We first introduce these datasets and corresponding implantation details. And then, we compare it with the state-of-the-arts by MPJPE. 

\subsection{Datasets and Implementation Details}
\begin{table*}[!t] 
\caption{Short and long-term prediction on CMU-mocap.}
\scriptsize
\begin{center}
\begin{tabular}{c|ccccc|ccccc|ccccc}
\hline
motion& \multicolumn{5}{c}{Basketball} & \multicolumn{5}{c}{Basketball Signal}& \multicolumn{5}{c}{Directing Traffic}\\
\hline
time (ms) & 80 &160 & 320 &400 &1000& 80 &160 & 320 &400 &1000& 80 &160 & 320 &400 &1000\\
\hline
LTD \cite{08}&14.0&25.4 &  49.6 &61.4&{106.1}   &3.5 &   6.1 & 11.7 &15.2 &   53.9& 7.4 & 15.1 &31.7 &   42.2 &152.4\\

LPJP \cite{32}&11.6&21.7 &  44.4 &57.3&{\bf90.9}   &2.6 &   4.9 & 12.7 &18.7 &   75.8& 6.2 & 12.7 &29.1 &   39.6 &149.1\\

TrajCNN \cite{20}&11.1&19.7   &43.9&   56.8& 114.1&   {\bf1.8} &3.5  &{\bf9.1}   &13.0 &{\bf49.6}  &{\bf5.5} & {\bf10.9}   &{\bf23.7} &{\bf31.3}&  {\bf105.9}\\
\hline
Ours&{\bf11.1}&{\bf19.5}   &{\bf42.8}& {\bf55.7}&  113.1&   1.9   &{\bf3.5}   &9.3  &{\bf13.0}  &57.5 &5.8 &   11.7  &25.6&33.4& 139.0\\
\hline
motion& \multicolumn{5}{c}{Jumping} & \multicolumn{5}{c}{Running}& \multicolumn{5}{c}{Soccer}\\
\hline
time (ms) & 80 &160 & 320 &400 &1000& 80 &160 & 320 &400 &1000& 80 &160 & 320 &400 &1000\\
\hline
LTD \cite{08}&16.9&  34.4 &   76.3  &96.8    &164.6   &25.5    &36.7&   39.3 &   39.9  &58.2 &11.3    &21.5    &44.2&   55.8 &   117.5\\

LPJP \cite{32}&12.9&27.6 &  73.5 &92.2&{176.6}   &23.5 & 34.2 & 35.2 &36.1 &   43.1& 9.2 & 18.4 &39.2 &   49.5 &{\bf93.9}\\

TrajCNN \cite{20}&12.2  &28.8&   {\bf72.1}&  94.6& 166.0 &17.1&   24.4  &28.4&   32.8& 49.2  &{\bf8.1}   &{\bf17.6}& 40.9  &51.3 &126.5\\
\hline
Ours&{\bf11.4} &{\bf28.0}& 72.7& {\bf94.1}&  {\bf155.3}  &{\bf16.4}& {\bf20.1}   &{\bf22.9}& {\bf27.6}&  {\bf41.9}   &8.6  &18.3&   {\bf39.1}   &{\bf48.4}  &{\bf103.6}\\
\hline

\end{tabular}
\begin{tabular}{c|ccccc|ccccc|ccccc}
\hline
motion& \multicolumn{5}{c}{Walking} & \multicolumn{5}{c}{Wash Window}& \multicolumn{5}{c}{Average}\\
 \cline{1-16}
time (ms) & 80 &160 & 320 &400 &1000& 80 &160 & 320 &400 &1000& 80 &160 & 320 &400 &1000\\
\hline
LTD \cite{08}&7.7 &  11.8 &   19.4 &   23.1 &   40.2  &5.9 &   11.9 &   30.3 &   40.0 &   79.3  &11.5 &  20.4 &   37.8 &   46.8 &   96.5\\

LPJP \cite{32}&6.7&10.7 &  21.7 &27.5&{\bf37.4}   &5.4 & 11.3 & 29.2 &39.6 &   79.1& 9.8 & 17.6 &35.7 &   45.1 &93.2\\

TrajCNN \cite{20}&6.5   &10.3&   19.4& 23.7& 41.6  &{\bf4.5}&{\bf 9.7}  &29.9&   41.5& 89.9  &8.3  &15.6&   33.4  &43.1 &92.8\\
\hline
Ours&{\bf5.9}  &{\bf9.0}&  {\bf17.4}&  {\bf21.1}&  {38.8}&  4.6   &10.0 &{\bf28.6}  &{\bf39.0}& {\bf73.1}   &{\bf8.2}&  {\bf15.1}   &{\bf32.3}& {\bf41.5}   &{\bf90.3}\\
\hline
\end{tabular}
\end{center}
\label{results_cmu}
\vspace{-2.5em}
\end{table*}
\textbf{H3.6M} \cite{24} is the most widely used benchmark for motion prediction. It involves 15 actions performed by professionals, and each human pose involves a 32-joint skeleton. Following \cite{08,20}, we compute the joint's 3D coordinates by applying forward kinematics and down-sample the motion sequence to 25 frames per second. To remove the global rotation, translation, and constant 3D coordinates of each human pose, there remain 22 joints. We test our method on subject 5 (S5).

\textbf{3DPW} \cite{25} The 3D Pose in the Wild dataset (3DPW) \cite{25} consists of challenging indoor and outdoor actions. The dataset consists of various activities such as shopping, doing sports, and hugging, including 60 sequences and more than 51k frames. For a fair comparison, we evaluate the whole test set.

\textbf{CMU-Mocap} The CMU mocap dataset mainly includes five categories. Be consistent with \cite{08,20}, we select 8 detailed actions: ``basketball'', ``basketball signal'', ``directing traffic'', ``jumping'', ``running'', ``soccer'', ``walking'' and ``washing window''.

\textbf{Network Setting}. We take three timescales: 3, 5, and 7 frames around the target frame in MTDE. The size of the high-level dimension $D_p$ is 32. We use 5 layers in the encoder and 4 layers in the decoder to get enough receptive field. The size of the temporal dimension is enlarged to 64. More details can be found in the supplementary material.

\textbf{Training}. All training is conducted on the Pytorch platform with one 2080Ti GPU. We use Adam \cite{26} optimizer with an initial learning rate of 0.0005. We use a weight decay of 0.96 and set the learning rate as 0.0001. The batch sizes are set to 16.

\subsection{Comparison with state-of-the-art}
Here we show the prediction performance for both short-term and long-term motion prediction on H3.6M, CMU Mocap, and 3DPW. We quantitatively evaluate various methods by the MPJPE between the generated motions and ground truths in 3D coordinates space. To be consistent with the literature\cite{20,08}, we report our results for short-term ($<$ 500ms) and long-term ($>$ 500ms) predictions. For all datasets, we are given 10 frames (400 milliseconds) to predict the future 10 frames (400 milliseconds) for short-term prediction and to predict the future 25 frames (1 second) for long-term prediction. More results can be found in the supplementary material.

\subsubsection{Results on H3.6M}
\textbf{Short-term motion prediction.} Table \ref{r_h36mshort} provides the short-term predictions on H3.6M for the 15 activities and the average results. Note that our method outperforms all the baselines on average and almost all motions. It demonstrates that our approach learns the general representation of different movements. Specifically, for those motions that need the upper body and lower body to cooperate, e.g., ``Walking dog'', ``Phoning'' and ``Sitting down'', our method outperforms the most, reflecting the efficacy of our proposed BA in joint relation modeling. Besides, the results on 320ms and 400ms increase most, which shows that our method is good at capturing temporal continuity compared with other methods. We also provide qualitative comparisons in Figure \ref{f_vis_all}. They further evidence that our predictions are closer to the ground truth than those of the above actions' baselines. More visualizations are included in the supplementary material.


\textbf{Long-term motion prediction.} In Table \ref{r_h36mlong}, we compare our results with those of the baselines for long-term prediction on H3.6M. Our method outperforms all the baselines on average. For long-term prediction, with the uncertainly of motion increasing, our method still obtains competitive performances on almost all motions. Especially in motions with more dynamics like “Walking Dog”, our method outperforms other competitors most. The observations demonstrate the advantages of our proposed dynamics representation and BA.

\subsubsection{Results on CMU-Mocap and 3DPW}

Table \ref{results_cmu} reports the MPJPE for short-term and long-term prediction on CMU-Mocap and Table \ref{r3dpw} reports the results on 3DPW. In essence, the conclusions remain unchanged: our method consistently outperforms the baselines for both short-term and long-term prediction with BA guidance.

\begin{table}[htb]
\caption{Short and long-term predictions on 3DPW. }
\scriptsize
\begin{center}
\centering
\begin{tabular}{c|ccccc}
\hline
{time (ms)} & 200 &400 & 600 &800 & 1000\\
 \hline
{LTD} \cite{08} & 36.0 &69.0 &91.0 &107.6 &118.6\\
\hline
{Ours} &{\bf 34.7} &{\bf 66.7} &{\bf 85.6}&{\bf 98.0} &{\bf 108.4}\\
 \hline
\end{tabular}
\end{center}
\label{r3dpw}
\vspace{-4.5em}
\end{table}

\section{Ablation study}
In this section, we conduct several ablation experiments on H3.6M to testify the effectiveness of different components in our proposed framework.

\subsection{Effectiveness of components of MTDE}

MTDE is designed mainly to get enriching dynamics information of raw input data.  Table \ref{ab_1} shows the results of experiments. The results of 320ms and 400ms increase significantly, which shows MTDE encodes more temporal information and offers more meaningful guidance for prediction, especially in the long time horizon.

\begin{table}[h]
\caption{Results of ablation experiments on MTDE}
\scriptsize
\begin{center}
\centering
\begin{tabular}{c|cccc}
\hline
MTDE&80&160&320&400\\
\hline
\XSolid & 9.8 &22.6 &48.0 &58.4 \\
\Checkmark&9.6 &22.0 &46.3 &57.0\\
\hline
\end{tabular}
\end{center}
\label{ab_1}
\vspace{-4em}
\end{table}
\subsection{Effectiveness of components of GCE}

GCE is designed mainly to model the global coordination of joints according to the nature of the human body to keep balance. It mainly has two components: Balance Attractor Unit (BAU) and Cosine Similarity Unit (CSU). To prove the effectiveness of CSU, we design an experiment with a common softmax function as a comparison. To prove the guidance of BA is useful, we also design an experiment without BAU. Here ``$Sim_c$'' and ``$Sim_s$'' represent the usage of cosine similarity and softmax respectively. ``BAU'' is the Balance Attractor Unit. Table \ref{ab_2} shows the results.
\begin{table}[h]
\caption{Results of ablation experiments on GCE}
\scriptsize
\begin{center}
\centering
\begin{tabular}{ccc|cccc}
\hline
$Sim_c$&$Sim_s$&BAU&80&160&320&400\\
\hline

\XSolid&\Checkmark &\Checkmark& 10.2 &23.4 &49.5&60.6\\
\XSolid&\XSolid &\XSolid & 10.1 &23.1 &49.2 &60.0 \\
\Checkmark&\XSolid &\XSolid& 9.7 &22.3 &47.4 &58.4 \\
\Checkmark&\XSolid&\Checkmark&9.6 &22.0 &46.3 &57.0\\

\hline
\end{tabular}
\end{center}
\label{ab_2}
\vspace{-4em}
\end{table}
We have the following observations: 

(1)   The BAU is essential for effective prediction, especially on long horizon. It demonstrates that the indirect BA offer useful guidance and this module extract meaningful global motion features.

(2)   The cosine similarity is better compared with the softmax function used in self-attention models. It arises from two aspects. First, it avoids violent differences in the softmax function because cosine similarity limits the value in $(-1,1)$. Second, it has the angle information to represent both orientation and intensity of correlation, while softmax only represents the intensity of correlation.

(3)   Methods with proposed GCE outperforms 0.5, 1.1, 2.9, 3.0 by the one without GCE for 80ms, 160ms, 320ms, 400ms, respectively. This proves the effectiveness of the GCE module.
\subsection{Effectiveness of LIE and AFFM}

In table \ref{ab_3}, the method with a single GCE outperforms the one with single LIE. This demonstrates that our proposed GCE is superior to those encodes local interactions of joints, which indicates the importance of our proposed BA. The improved performance due to fusing these two paths proves that these two paths are complementary. 

AAFM improves the results by 0.4 on average. It reflects that the channel attention enhances the whole performance. Besides, it increases slowly compared with the introduction of GCE and LIE, which reflects that our model's improvement mainly benefits from the design of GCE and LIE.

\begin{table}[h]
\scriptsize
\begin{center}
\centering
\caption{Results of ablation experiments on LIE and AFFM}

\begin{tabular}{ccc|cccc}
\hline
$GCE$&$LIE$&AFFM&80&160&320&400\\
\hline
\Checkmark&\XSolid &\Checkmark&  9.7 &22.6 &48.3 &58.9\\

\XSolid&\Checkmark &\Checkmark& 10.1 &23.1 &49.2 &60.0 \\
\Checkmark&\Checkmark &\XSolid & 9.6 &22.4 &46.8 &57.4 \\

\Checkmark&\Checkmark &\Checkmark&9.6 &22.0 &46.3 &57.0\\

\hline
\end{tabular}
\end{center}
\label{ab_3}
\vspace{-4.5em}
\end{table}

\section{Conclusion}
In this paper, we have proposed a simple yet effective framework referred to as Attractor-Guided Neural Network to model spatiotemporal features for skeleton-based human motion prediction. We extract the dynamic representation of raw skeleton data from a MTDE for effective prediction. To exploit richer joint relation, we propose an AJRE module to better leverage joint relation, including GCE and LIE. The former presents global coordination of all joints and later encodes local interactions between joint pairs. With those two fine-grained features introduced, our proposed method achieves state-of-the-art results on three benchmark datasets.

{\small
\bibliographystyle{ieee_fullname}
\bibliography{egbib}
}

\end{document}